# CCi-YOLOv8n: Enhanced Fire Detection with CARAFE and Context-Guided Modules

Kunwei Lv[1][0009-0002-5873-865X], Ruobing Wu[1][0009-0006-0062-9994], Suyang Chen[2], and Ping Lan✉[0009-0006-4245-4010]

[1] Tibet University, Lhasa, China
`lkw872473172@163.com`
`lanping@utibet.edu.cn`

[2] The Grove School of Engineering - Computer Science, City University of New York: City College, New York, 10031, United States
`schen092@citymail.cuny.edu`

**Abstract.** Fire incidents in urban and forested areas pose serious threats, underscoring the need for more effective detection technologies. In order to overcome these difficulties, we present CCi-YOLOv8n, an enhanced YOLOv8 model with targeted improvements to detect small fires and smoke. The model integrates the CARAFE upsampling operator and a context-guided module to reduce information loss during upsampling and downsampling, thereby retaining richer feature representations. Furthermore, an inverted residual mobile block enhanced C2f module captures small targets and fine smoke patterns, a critical improvement over the original model's detection capacity. For validation, we introduce Web-Fire, a dataset curated for fire and smoke detection in a range of real-world situations. The experimental results indicate that CCi-YOLOv8n outperforms YOLOv8n in detection precision, confirming its effectiveness for robust fire detection tasks.

## 1   Introduction

Fire outbreaks, whether in urban centers or remote wilderness, can cause devastating losses due to delayed responses. Manual detection methods are slow and labor intensive and require constant human oversight. In contrast, automatic fire detection enables real-time alerts that can curb fire spread and reduce casualties. Recent advances include deep learning systems for wildfire detection [1,2], neural fuzzy systems for complex forest environments [3], and enhanced YOLOv8 models for urban fires [4,5]. However, most models are environment-specific. Combining Himawari-8 satellite images with convolutional neural networks [6,7] improved active fire detection accuracy, although real-time capability remains a challenge. A recent MobileNetV3-YOLOv4 model for embedded devices [8] provides faster detection, although at the cost of some loss of feature details, which impacts the overall detection quality. Given the limitations of previous work, this study adopts YOLOv8n as the baseline model, As a result of its superior behavior in the field of object detection. The original model utilizes a simplistic upsampling method in the neck, which doubles the height (H) and width (W) while keeping the number of channels unchanged. In the backbone's downsampling operations, standard convolution is used, which halves H and W and doubles the number of channels, thereby resulting insignificant feature loss during sampling. Although the backbone's C2f module enhances the original C3 module by incorporating a split operation and leveraging residual connections for optimization, the bottleneck component of the original model employs a relatively basic feature extraction approach, limiting its capacity for comprehensive feature extraction. The original detection model still exhibits some shortcomings in terms of detection and false detection. The heat maps for both YOLOv8n and the enhanced CCi-YOLOv8n model during the detection process are presented in Fig. 1.

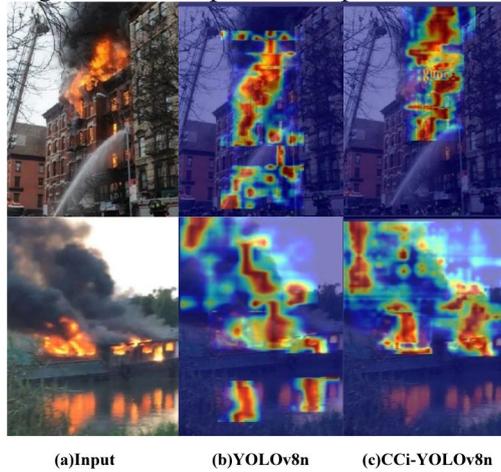

(a)Input          (b)YOLOv8n          (c)CCi-YOLOv8n

**Fig. 1.** The heatmap generated from the input image has poor detection results, where the red box represents the part that the heatmap does not detect, and the blue box represents the part that is falsely detected.

Therefore, we propose an improved model, CCi-YOLOv8n, to address these deficiencies. The proposed improvements aim to mitigate the feature loss issues during sampling and enhance feature extraction throughout the network, thereby improving overall model performance in object detection tasks. The model provides a universal solution with improved real-time performance, data retention, and minimal computational load. The Web-Fire dataset, which we developed, contains fire and smoke instances from both urban and wild settings, and is complemented by the D-Fire dataset [9], covering complex fire scenes. By incorporating novel sampling algorithms and an iRMB to enhance the original C2f, our model addresses significant false detection and detection omission rates, along with improved bounding box accuracy in diverse fire and smoke scenarios. The key contributions of this paper are listed as follows:

• This paper introduce the CARAFE upsampling operator and the Context Guided downsampling module to optimize information retention during the sampling process, ensuring a more comprehensive understanding of contextual content to facilitate the complete transmission of features.

- This paper utilizes the iRMB module to improve detection accuracy for small targets and smoke. This improves the multi-scale feature extraction capability of the original C2f module. To ensure speed, This paper instegrates depthwise convolution and reevaluate lightweight CNN architecture from the perspective of efficient unification of IRB and Transformer components. Balancing parameters while achieving superior detection performance over comparable attention-based models.

## 2 Related Work

### 2.1 Sample

In object detection [10–12], sampling primarily serves to balance the class distribution by effectively selecting positive and negative samples, thus enhancing model accuracy and training efficiency. Imbalances between positive and negative samples can degrade performance, making sampling strategies essential to optimize detection capabilities. Effective sampling allows models to focus on critical samples for accurate classification and localization. Dysample [13] introduces an ultralightweight dynamic upsampler that is resource-efficient and easily imple-mented in PyTorch. Building on this, MFDS-DETR [14] introduces HS-FPN, which further evolves into HSPAN. Downsampling, a common technique for reducing spatial dimensions, is refined with SPD-Conv [15], a new CNN block replacing strided convolution and pooling layers. YOLOv9 [16] presents "adown", a lightweight downsampling approach that optimizes detection accuracy and efficiency. Each method uniquely improves feature retention during sampling. In contrast, this paper introduces the CARAFE [17] and CGD [18] modules, which enhance model accuracy by capturing contextual features. Comparative experiments will further demonstrate the effectiveness of these proposed sampling improvements. Recently, diffusion models have achieved remarkable success in generative tasks, including cross-modal and pose-guided generation. Shen et al. [19] propose MCDM for long-term talking face generation with enhancedidentity consistency. IMAGPose [20] unifies pose-guided person generation under aconditional diffusion framework. For virtual try-on, IMAGDressing-v1 [21] provides fine-grained control of garment transfer. Progressive conditional diffusion has also been applied to boost fidelity in pose-guided synthesis [22], and story visualization tasks [23] demonstrate the flexibility and potential of diffusion-based generation. Inspired by these advances, our method incorporates transformer-based attention into a diffusion-style sampling structure for robust and context-aware detection.

### 2.2 Transformer

In object detection, Transformers be essential in boosting feature extraction and information aggregation via a powerful self-attention mechanism. This architecture effectively captures global dependencies among objects, thus improving detection accuracy. Unlike traditional Convolutional Neural Networks (CNNs) [24–26], Transformer model identifies relationships across the entire feature map, facilitating a more comprehensive depiction of object features. Furthermore, Transformers reduce the reliance on anchor boxes and simplify network design, contributing to a more efficient and adaptable detection pipeline. Initially, object detection relied heavily on CNNs for feature extraction and localization. However, in 2020, the introduction of DETR (Detection Transformer) [27] marked a significant shift, bringing transformers into detection through an end-to-end approach that simplifies complex anchor and post-processing steps. DETR demonstrated the Transformer's strengths in modeling long-range dependencies and integrating global features, proving highly effective in complex scenes.Following DETR, models like Deformable DETR [28] addressed DETR's slow training and convergence by incorporating a deformable attention mechanism, significantly boosting detection speed and accuracy. Additionally, models such as Swin Transformer [29] combine convolution and Transformer advantages, capturing multi-scale features and improving performance on small objects and complex backgrounds. However, like larger CNNs, Transformers increase model parameters, which can slow down detection. To address this, we introduce the iRMB [30], combining CNN and Transformer elements to improve the C2f module in the original network. This approach leverages both architectures while maintaining detection speed.

## 3   Proposed Method

### 3.1   Overview

To address the high rates of false negatives and false positives in YOLOv8n[31] for fire detection tasks, this paper introduces CARAFE and CGD sampling modules into original model. Additionally, the existing C2f modules in both sections were enhanced by incorporating iRMB, resulting in the new C2f_iRMB layer. In the Backbone section, the original down-sampling convolutional layer was replaced with the CGD down-sampling module, as shown in part ①. In the Neck section, the original Upsample layer was replaced by the CARAFE module, corresponding to part ②. Moreover, the C2f layers in both sections were enhanced by integrating the iRMB module, leading to the new C2f_iRMB layer, where n indicates the number of iRMB modules that can be used, as shown in part ③.The improved structures as shown in Fig. 2.

### 3.2   Content-aware Reassembly of Features Upsampling Module

The CARAFE predicts the reassembly kernel based on the content at each target position, and utilizes the predicted kernel to reconstruct features. Kernel Prediction first reduces the input channels through a $1 \times 1$ convolution to minimize computational cost, followed by a $k \times k$ convolution layer to predict an enlarged upsampling kernel, representing a larger receptive field. The Content Encoder acts as a $k \times k$ upsampling kernel. Finally, SoftMax is used to normalize the predicted up-sampling result. Content-Aware Reassembly maps the output back to the input and per- forms a dot product operation between the $k \times k$ area centered around each point and the predicted result to get the output.

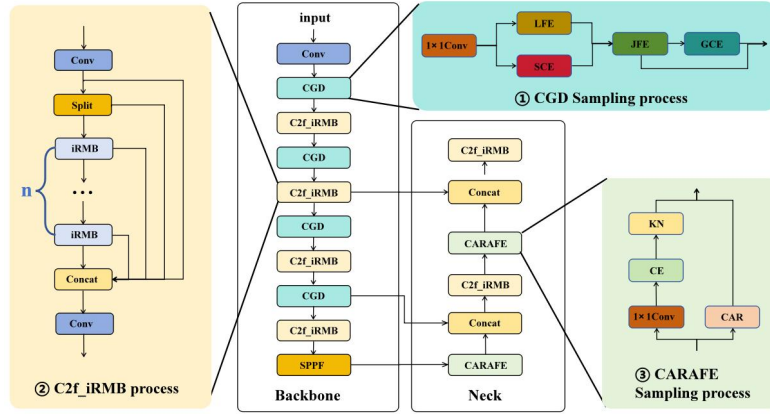

**Fig. 2.** Illustrates the modifications made to CCi-YOLOv8n

Given a feature map $X$ of size $H \times W$, the upsampling process produces a new feature map $X^*$ of size $H^* \times W^*$. In the output species there is $l' = (i', j')$ corresponding to the original position $l=(i, j)$ in the input. Here, $N(X_l, k)$ represents the $N(X_l, k)$ sub-region centered on position $l$ in $X$. At first, the kernel prediction module $W$ predicts the kernel $X_l$ for each position $l'$ based on the neighbors of $X_l$, as in Eq.(1). The reassembly step is formalized in Eq.(2), where $\varphi$ responsible for reassembling the neighboring components of $X_l$ with the kernel $W_l$.

$$W_l = \psi(N(X_l, k_{encoder})). \tag{1}$$

$$X_l = \phi(N(X_l, k), W_l). \tag{2}$$

The introduction of the CARAFE module helps the model to retain more complete feature information in the Neck section, thereby improving the accuracy of bounding box predictions. The detailed sampling process of the module is illustrated in Fig. 3.

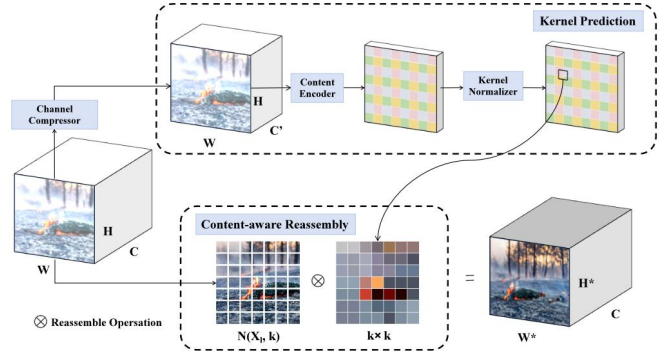

**Fig. 3.** The overall framework of CARAFE, CARAFE is composed of kernel prediction module and content-aware reassembly module. Feature maps of size $C \times H \times W$ is double upsampled.

### 3.3    Content-aware Reassembly of Features Up-sampling Module

The CGD module operates by first compressing the input via a 1×1 convolution. Then pass through Local Feature Extractor (LFE), which uses standard convolution layers to learn local features, and the Surrounding Context Extractor (SCE), which utilizes dilated convolution to capture larger receptive field context. The Joint Feature Extractor (JFE) then fuses local features with surrounding context using Batch Normalization (BN) and ReLU operations. The Global Context Extractor (GCE) performs global average pooling to aggregate the context and by two fully connected (FC) layers to weight the features, enhancing the useful parts while suppressing the irrelevant parts. The LFE and SCE work together to ensure that the model not only understands the information of each pixel or local area but also understands the relationships between these areas in the overall context. This detail and local variation overview information is fundamental for the precise classification of each pixel in the image, especially for tasks requiring fine-grained predictions, such as distinguishing different objects and surfaces in complex scenes. This design helps CGD to operate efficiently even in resource-limited environments preserving high accuracy and real-time efficiency performance at the same time.

By using dilated convolutions, the SCE enlarges the receptive field, enabling the model to observe a larger region rather than just focusing on local details. By combining LFE, additional information is provided, helping the model better understand complex scenes. The role of JFE is to integrate features extracted by LFE and SCE, preserving local details while incorporating broader information of context. JFE empowers the network to integrate both local and contextual information, resulting in improved semantic segmentation accuracy. GCE produces a global feature vector through global average pooling, which captures the average response for each channel in the input image. The global feature vector is then further processed by a multi-layer perceptron (MLP), The extracted global context is combined with the joint features via a scale layer, which adjusts the joint channel-level features, emphasizing the useful features and suppressing the unimportant ones. GCE generates the extracted global contextual information based on the input image, allowing the network to generate customized global context for different images. The introduction of the CGD module helps the model better integrate different local features within the target, enabling more complete feature trans- mission and reducing the false detection rate while improving the model's testing accuracy. Expanded Window Multi-Head Self-Attention (EW-MHSA): EW-MHSA reduces the calculation cost of the traditional self-attention mechanism by executing multi-head self-attention computations. The process is described by Eq.(3),Eq.(4)and Eq.(5).

$$Q = K = X. \qquad (3)$$

$$V = Expand(X). \qquad (4)$$

$$Attention(Q, K, V) = softmax(\frac{QK^T}{\sqrt{d_k}})V. \qquad (5)$$

3. Feed-Forward Network (FFN): After EW-MHSA, the feature map is further transformed by the feed-forward network. Its computation is represented by Eq.(6).

$$X_{ffn} = FFN(X_{att}). \tag{6}$$

4. Residual Connections: To facilitate efficient gradient flow and avoid gradient vanishing, residual connections are added to iRMB. These connections ensure that the output retains input information, assisting in training deeper networks. Residual connections are employed to generate the final output, it described by Eq.(7)

$$Y = X + X_{ffn}. \tag{7}$$

By combining DW-Conv and EW-MHSA, iRMB strikes an effective balance between capturing local and global features. The attention mechanism within the local window making this design more feasible for practical applications. Improving the original C2f module using this block enhances the model's ability to detect small targets and reduces the missed detection rate.The sampling process of the CGD module is shown in detail in Fig. 4.

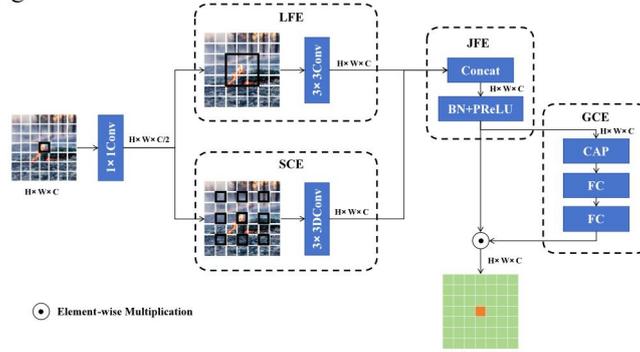

**Fig. 4.** The overall framework of Context Guided Downsampling Module.

### 3.4  Inverted Residual Mobile block Module

The original bottleneck block had limited multi-scale feature extraction capabilities, but adding multi-head attention mechanisms alone would significantly increase the model's computational cost. Therefore, this paper propose to combine the inverted residual movable block (iRMB) with the bottleneck block in C2f to boost efficiency on models. The iRMB combines DW-Conv and EW-MHSA to effectively balance local and global feature extraction, Specifically, DW-Conv is used to efficiently extract spatial features with low computational complexity. DW-Conv operates on each channel individually, significantly reducing the number of parameters and computation. Its computation is defined by Eq.(8).The overall framework of the iRMB module is shown in Fig. 5.

$$X_{dw} = DW - Conv(X). \tag{8}$$

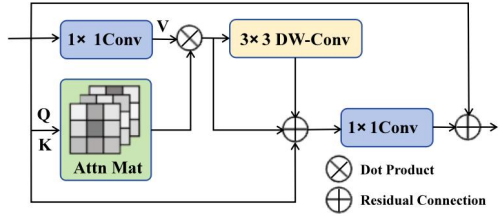

**Fig. 5.** The iRMB module first extracts value (V) features through a 1x1 convolution, followed by generating the attention matrix (Attn Mat) using query (Q) and key (K). The attention matrix is then multiplied with the value features, and the weighted features are processed by a 3x3 depth-wise convolution (DW-Conv), then added to the input. Subsequently, another 1x1 convolution is applied to integrate the features.

## 4　　Experiment and Analysis

The experimental setup of this study includes the following: CPU: INTEL-XEON 6230, GPU: NVIDIA GeForce RTX 4090, system environment: programming language: Python 3.9, accelerated environment: CUDA 12.2, PyTorch version: 2.0.0. The datasets used in the experiment include a custom fire dataset named "Web-Fire" and an open dataset named "D-Fire."

### 4.1　　Inverted Residual Mobile block Module

**D-Fire.** D-Fire dataset is a dataset developed by Gaia researchers specifically for fire and smoke detection, containing over 21,000 images divided into four classes: fire only (1,164 images), smoke only (5,867 images), both fire and smoke (4,658 images), and neither fire nor smoke (9,838 images).

**Web-Fire.** The custom fire dataset contains 11,646 images, divided into 10,481 training images and 1,165 test images with a ratio of 9:1, ensuring the complexity and completeness of the dataset, allowing the model to accurately detect various fire scenarios.

### 4.2　　Evaluation Metrics

The evaluation indicators for experiments in this paper include Precision, Recall, and mAP. where precision is how many of the samples that the model predicts to be positive are correct, and recall is how many of the samples that are truly positive are correctly predicted to be positive. which represent the average accuracy of IoU between 0.5 and 0.5-0.95, respectively.

### 4.3　　Comparison with State-of-the-art Methods

Table 1. present a comparison between CCi-YOLOv8n and other models on the Web-Fire and D-Fire datasets. In addition to the four indicators mentioned earlier, parameter and Flops are also introduced to better reflect the comparison between models. The results reveal that the CCi-YOLOv8n model achieves superior target detection performance compared to others. Specifically, CCi-YOLOv8n attained 72.0% and 41.0% for the two metrics on the Web-Fire dataset, and 78.5% and 46.6% on the D-Fire dataset.Our model and other versions of YOLO with similar

model sizes, including YOLOv9, YOLOv10, and YOLOv11, performed better in comparison. Since the new version of YOLO is lighter than YOLOv8, the parameters and Flops of the model increased albeit slightly , but the performance of the model can make up for this shortcoming. Additionally, CCi-YOLOv8n has fewer parameters and FLOPs, achieving a substantial reduction in computational requirements compared to earlier models while maintaining high precision.

**Table 1.** Comparisons on Web-Fire and D-Fire.

| ID | Method | Params (M) | Flops (G) | Web-fire | | | | D-fire | | | |
|---|---|---|---|---|---|---|---|---|---|---|---|
| | | | | Precision(%) | Recall(%) | mAP 50(%) | mAP50:95(%) | Precision(%) | Recall(%) | mAP 50(%) | mAP50:95(%) |
| A | SSD[32] | 34.3 | 31.4 | 48.6 | 38.9 | 36.4 | 12.5 | 57.9 | 52.0 | 51.7 | 20.5 |
| B | FCOS[33] | 32.2 | 184 | 66.6 | 65.9 | 67.3 | 32.9 | 68.6 | 63.5 | 65.2 | 29.7 |
| C | Faster-r-cnn[34] | 33.8 | 130 | 61.7 | 61.7 | 61.2 | 27.5 | 64.3 | 56.3 | 58.8 | 24.7 |
| D | Retinanet[35] | 36.5 | 210 | 69.0 | 63.9 | 66.3 | 32.8 | 72.8 | 63.6 | 69.7 | 33.1 |
| E | YOLOv5s[36] | 7.0 | 16.0 | 66.9 | 65.5 | 66.9 | 33.7 | 78.2 | 71.4 | 76.7 | 43.0 |
| F | YOLOv7-tiny[37] | 6.0 | 13.1 | 65.7 | 66.9 | 68.0 | 32.7 | 75.4 | 70.2 | 76.3 | 40.6 |
| G | YOLOv8n | **2.0** | 7.7 | 66.8 | 69.0 | 69.8 | 40.6 | 78.1 | 70.2 | 77.3 | 45.6 |
| H | YOLOv9t[38] | 2.6 | 8.2 | 67.3 | 66.4 | 68.5 | 39.7 | 76.8 | 67.1 | 74.7 | 44.1 |
| I | YOLOv10n[39] | 8.0 | 24.4 | 68.5 | 64.9 | 68.7 | 40.2 | 78.6 | 69.4 | 77.5 | 46.2 |
| J | YOLOv10s | 2.6 | **6.4** | 65.7 | 66.9 | 68.0 | 32.7 | 77.9 | 69.3 | 76.8 | 45.1 |
| K | YOLOv11n | 3.2 | 8.2 | 63.6 | 68.7 | 68.4 | 38.2 | 79.3 | 69.6 | 75.4 | 43.3 |
| L | CCi-YOLOv8n(ours) | 3.8 | 8.4 | **69.2** | **68.8** | **72.0** | **41.0** | **79.5** | **71.5** | **78.5** | **46.6** |

### 4.4 Ablation Studies and Analysis

Studies prove that the proposed CCi-YOLOv8n model is superior to many state-of-the-art object detection models. In the subsequent sections, the CCi-YOLOv8n method is analyzed in detail across three dimensions to explore the logic behind its superior performance. (1) Role of CARAFE upsampling.Compared with other upsampling modules including dysample and HSPAN. ours performs the best on the web fit dataset and dfire dataset. (2) Influence of CGD downsampling.In terms of performance in the downsampling section, our module has also shown the best performance along with Adown and SPDConv. (3) Impact of iRMB-C2f.Our module has been compared with recently proposed PKI and RVB methods and has shown the best performance.The comparison results presented in Fig. 6.

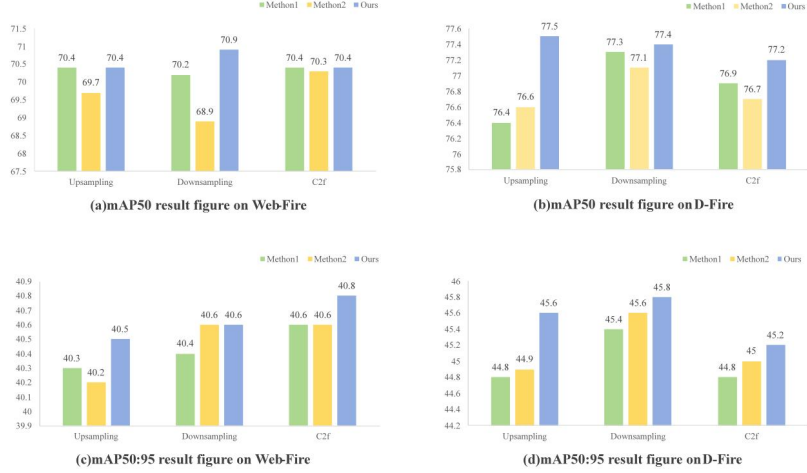

**Fig. 6.** Method1 of the upsampling part is dysample,and method2 is HSPAN. Method1 of the downsampling part is adown, and method2 is SPDConv.Method1 of the C 2f part is RVB[40],and method2 is PKI[41],The modules used in this paper have good performance in each part.

The experimental results indicate that when CARAFE and CGD modules were introduced individually, significant improvements in precision and mAP were observed, especially in mAP50. When both CARAFE and CGD were introduced together, the model's overall performance was optimal, demonstrating the synergy between CARAFE and CGD modules. Consequently, leveraging both of these modules can notably improve the fire and smoke detection system's performance, providing more accurate and reliable results, the results presented in Table 2.

**Table 2.** The results(%) ablation tests on Web-Fire and D-Fire.

| ID | CARAFE | CGD | iRMB | Web-fire | | | | D-fire | | | |
|---|---|---|---|---|---|---|---|---|---|---|---|
| | | | | Precision(%) | Recall(%) | mAP50(%) | mAP50:95(%) | Precision(%) | Recall(%) | mAP50(%) | mAP50:95(%) |
| 1 | | | | 68.7 | 68.4 | 38.4 | 79.3 | 69.6 | 75.4 | 43.3 | 20.5 |
| 2 | √ | | | 69.0 | 65.8 | 70.3 | 40.0 | 79.0 | 71.1 | 77.8 | 45.9 |
| 3 | √ | √ | | 69.1 | 68.8 | 70.9 | 40.0 | 78.6 | 71.7 | 78.2 | 46.4 |
| 4 | √ | √ | √ | 69.2 | 68.8 | 72.0 | 41.0 | 79.5 | 71.7 | 78.5 | 46.6 |

### 4.5  Visualization

To visually demonstrate the effectiveness of the our model, examples of detection on problematic data from the original model were selected, as shown in Fig. 7. (a) shows cases of missed detection; (b) illustrates false detection cases, and (c) demonstrates incomplete bounding boxes. The experiments indicate that the improved model not only detects objects that were previously undetected but also reduces the false detection rate and provides more complete target detection, validating the advancement of the improved model with increased overall accuracy.

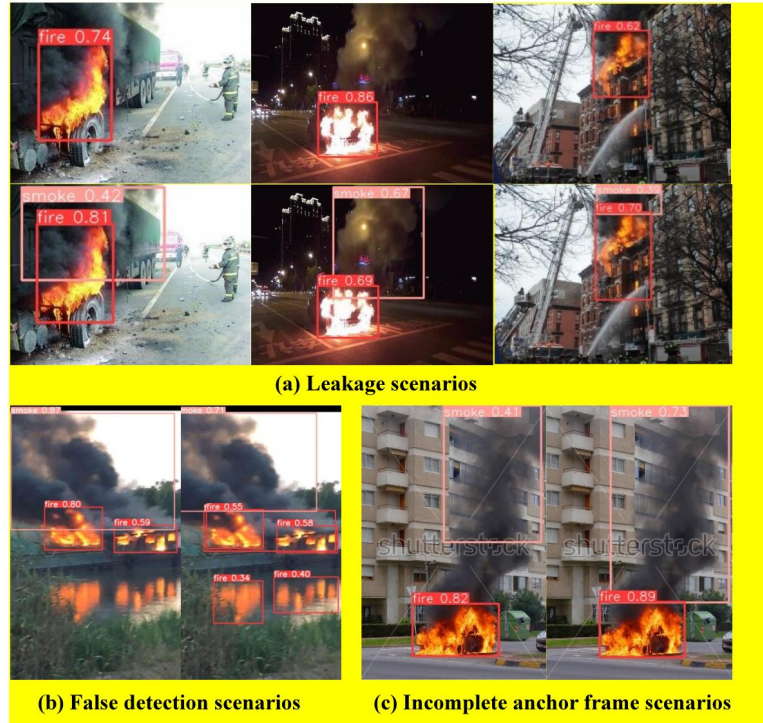

**Fig. 7.** A visual comparison of the original model with our model in three scenarios.

### 4.6 Conclusion

In this study, a fire dataset named "Web-Fire" was proposed, alongside an improved model CCi-YOLOv8n. Extensive experiments were conducted on both the custom-made Web-Fire and the publicly available D-Fire dataset. The results proved that the improved model enhances accuracy while incurring minimal additional computational overhead. The proposed model showed better performances across different datasets, highlighting its versatility. Future research will concentrate on boosting the model's generalizability and computational speed for wider practical use.

**Acknowledgments.** This study was funded by Lhasa science and technology plan project (grant number LSKJ202405).